\documentclass[journal,transmag]{IEEEtran}

\usepackage{epsfig,latexsym}
\usepackage{float}
\usepackage{indentfirst}
\usepackage{amsmath}
\usepackage{amssymb}
\usepackage{times}
\usepackage{psfrag}
\usepackage{setspace}
\usepackage{cite}
\usepackage{textcomp}
\usepackage{stfloats}
\usepackage{url}
\usepackage{subfigure}
\usepackage{booktabs}
\usepackage{graphicx}
\usepackage{bm}
\usepackage[ruled,vlined]{algorithm2e}
\usepackage{color}
\usepackage{mathrsfs}

\author{Yuantong Zhang,
	Huairui Wang,
	Han Zhu,
	Zhenzhong Chen,~\IEEEmembership{Senior~Member,~IEEE}
	\thanks{This work was supported in part by the National Natural Science Foundation of China under Contract 62036005 and the Special Fund of Hubei Luojia Laboratory. (Corresponding author: Zhenzhong Chen.)
		The authors are with the School of Remote Sensing and Information
		Engineering, Wuhan University, Wuhan, Hubei 430072, China . Z. Chen  is also with the Hubei Luojia Laboratory, Hubei 430079, China.
	}
}

\begin{document}
	\title{ Optical Flow Reusing for High-Efficiency Space-Time Video Super Resolution 
	}
	
	\maketitle
\markboth{IEEE TRANSACTIONS ON CIRCUITS AND SYSTEMS FOR VIDEO TECHNOLOGY}
{Y. Zhang \MakeLowercase{\textit{et al.}}: Optical Flow Reusing for High-Efficiency Space-Time Video Super Resolution}

\begin{abstract}
In this paper, we consider the task of space-time video super-resolution (ST-VSR), which can increase the spatial resolution and frame rate for a given video simultaneously. 
Despite the remarkable progress of recent methods, most of them still suffer from high computational costs and inefficient long-range information usage. To alleviate these problems, we propose a Bidirectional Recurrence Network (BRN) with the optical-flow-reuse strategy to better use temporal knowledge from long-range neighboring frames for high-efficiency reconstruction. Specifically, an 
efficient and memory-saving multi-frame motion utilization strategy is proposed by reusing the intermediate flow of adjacent frames, which considerably reduces the computation burden of frame alignment compared with traditional LSTM-based designs. In addition, the proposed hidden state in BRN is updated by the reused optical flow and refined by the Feature Refinement Module (FRM) for further optimization.
Moreover, by utilizing intermediate flow estimation, the proposed method can inference non-linear motion and restore details better.
Extensive experiments demonstrate that our optical-flow-reuse-based bidirectional recurrent network (OFR-BRN) is superior to state-of-the-art methods in accuracy and efficiency.  

\end{abstract}
\begin{IEEEkeywords}
  Video Super-Resolution, Video Frame Interpolation, Optical Flow
\end{IEEEkeywords}
\section{Introduction}
\IEEEPARstart{W}ITH  the development of video technologies and products, the public's pursuit of audiovisual experience continues to increase significantly. High-resolution  and high-frame-rate televisions are becoming more and more popular. However, many video sources are captured with low frame rate and resolution. Therefore, improving the resolution and frame rate of such videos has aroused considerable interests from researchers.  Heyden \emph{et al.}\cite{2002} first introduce the concept of Space-Time Video Super-Resolution (ST-VSR), which aims to transform a low spatial resolution video with a low frame rate into a video with higher spatial resolutions and temporal resolutions simultaneously.
Some  traditional methods \cite{2011Space,2013Adaptive} adopt manually designed feature extractors\cite{2006A,2003Super} to extract features and then perform pixel regularization to exploit spatiotemporal information. 
\begin{figure}[t]
	\centering
	\includegraphics[width=8.5cm]{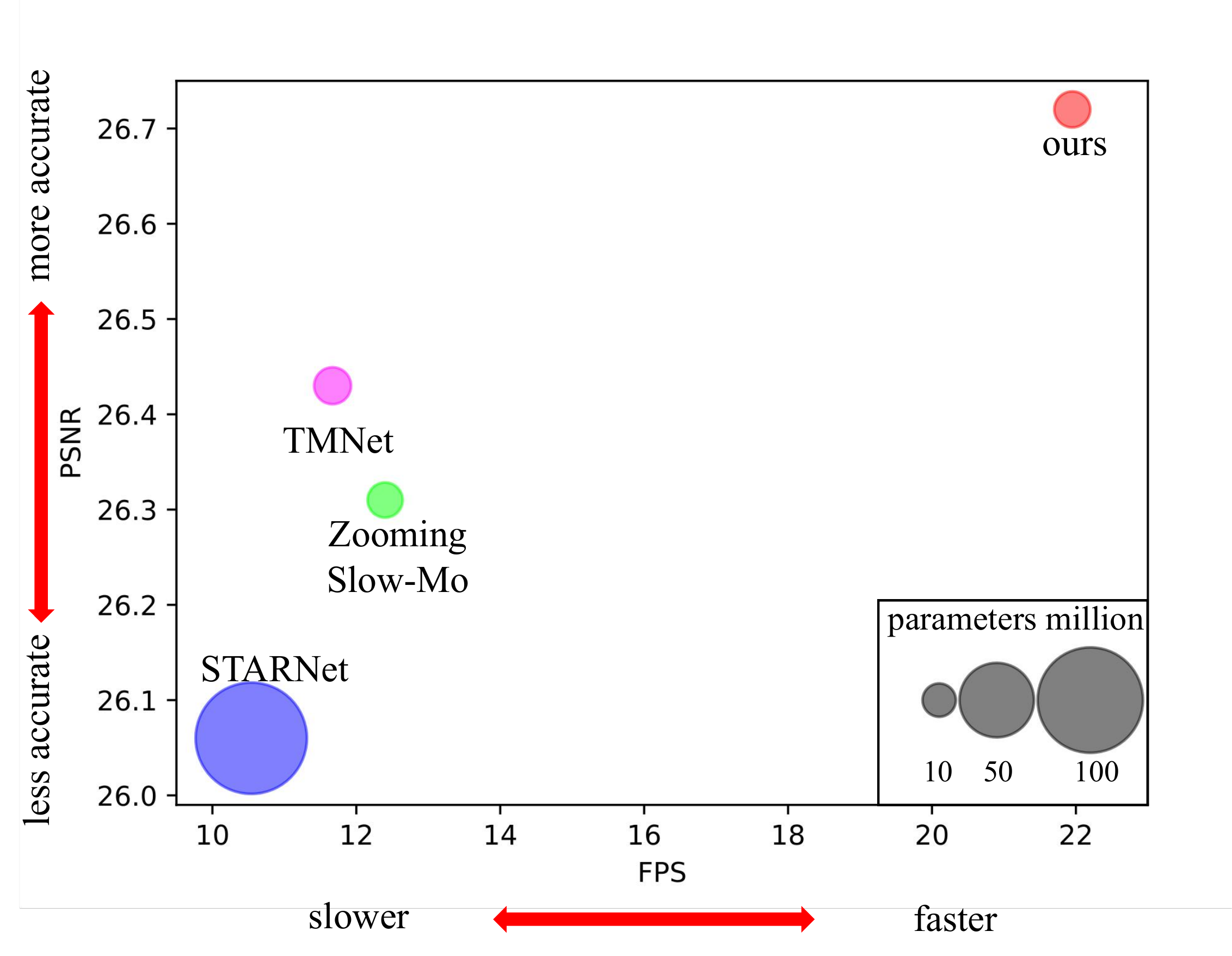}
	\caption{Comparison of accuracy (PSNR) and speed (FPS)\label{fig:1} of different
		methods on Vid4\cite{5995614} dataset. Our method is faster and more accurate than other state-of-the-art methods, such as Zooming Slow-Mo\cite{2021Zooming}, STARnet\cite{2020Space}, and TMNet,\cite{xu2021temporal} while keeping a relatively small amount of parameters. The inference speed is achieved on Vid4\cite{5995614} dataset when performing space $4 \times$ and time $2 \times$ super-resolution using one Nvidia 1080ti GPU.} 
	\label{fig:graph}
\end{figure}
These approaches are founded on complicated assumptions and hand-crafted priors. 
As a result, the reconstruction results are often unsatisfactory when modeling complex and diverse scenes.
With the development of deep learning, many video super-resolution methods\cite{2019EDVR,2019Recurrent,2018Deep } and video interpolation methods\cite{2017Video,DBLP:journals/corr/abs-1711-09078,2020Video} based on convolutional neural network (CNN) have made significant progress.  In fact, space-time video super-resolution can be accomplished in two steps by performing the video frame interpolation (VFI) method and the video super-resolution (VSR) method independently. However, the two-step scheme undoubtedly ignores the correlation of temporal and spatial information. In addition, since the alignment of frames or features is critical for both VFI and VSR, the two-step scheme and the joint training scheme inevitably need to perform alignment twice.  This will slow down the processing speed and introduce a large amount of parameter redundancy.

In the past few years, benefiting from the research advance of VFI and VSR, ST-VSR has also drawn increasing attention. Some recent work has been proposed to address the issues with two-step methods, aiming to achieve spatiotemporal reconstruction simultaneously.  STARnet\cite{2020Space} first estimates the optical flow of two adjacent frames, performs feature warping to interpolate the intermediate frame, and finally reconstructs high-resolution frames.
However, this method only considers the information of two adjacent frames but fails to leverage the information across misaligned long-range frames.
Some other methods use the ConvLSTM\cite{2015Convolutional} structure to restore high-resolution frames and introduce more complementary information from long-range neighboring frames.  Xiang \emph{et al.}\cite{2021Zooming} propose a deformable ConvLSTM backbone and perform ST-VSR in the feature space. Compared with STARnet, this method can handle relatively longer input video sequences with less computation and memory. However, its heavy reliance on deformable convolution\cite{Dai_2017_ICCV} for alignment and redundant motion estimation, on the other hand, may result in severe speed degradation. In addition, since its kernel-based motion estimation module is learned in an unsupervised manner, the predicted frames may suffer from blur when fast-moving objects exist.
\par
To design an efficient ST-VSR network, we carefully consider the relationship between VFI and VSR task. Because both VFI and VSR need to utilize information from multiple frames and perform motion estimation, we should find a way to connect the two tasks better 
and avoid redundant motion information estimation.  The key to both tasks is the alignment operation. For VSR, the information from other frames needs to be aligned to the current candidate frame. For VFI, data from both sides are employed to generate non-existent intermediate frames. Therefore, an appropriate representation strategy of motion information is needed to bridge the two tasks.  Given the linear additivity of optical flow, we propose a flow reorganization method  to generate the motion information for VSR and VFI simultaneously. The benefits of the proposed approach are twofold. First, directly estimating the optical flow of the intermediate frame can better approximate the non-linear and asymmetric motion. Second, with the aid of the optical-flow-reuse strategy, the generated intermediate optical flow can also be applied to the motion compensation of pre-existing frames. To summarize, a straightforward yet effective ST-VSR model is proposed with the optical-flow-reuse strategy for high-efficiency reconstruction. Specifically, We first inherited the excellent design of the bidirectional recurrence to ensure that all frames could utilize more information from the whole sequence. 
Then, the optical-flow-reuse strategy is adopted to renew and align the continuously updated state. Lastly, we employ the Feature Refinement Module (FRM) to optimize the hidden state and suppress possible noise before the final reconstruction. 
Our contributions are summarized as:
\begin{itemize}
	\item An optical-flow-reuse-based bidirectional recurrence network is proposed for effective and efficient long-range temporal knowledge leverage. The proposed scheme estimates the intermediate flow for multiple adjacent frames and then reuses them with flow reorganization. Benefiting from the flow reorganization mechanism, our model can better estimate the asymmetric motion and significantly increase the inference speed while consuming less memory. \\
	\item A Feature Refinement Module (FRM) is proposed to reduce possible errors invoked by inaccurate flow warping and further improve the reconstruction performance by refining the continuously updated hidden states.  \\
	\item By integrating the optical-flow-reuse strategy and FRM into the bidirectional recurrence network, our proposed structure can leverage multi-frame motion more efficiently than existing LSTM-based algorithms and significantly outperform state-of-the-art methods. (Fig.~\ref{fig:1}) 
	
\end{itemize}
The remainder of the paper is organized as follows:
Section II reviews the background and related work. 
Details of our proposed flow-reuse-based bidirectional network are given in Section III.
Experiments and analyses are provided in Section IV.  Finally, the paper is concluded in Section V.

\section{related Work}
\label{sec:intro}

\smallskip
\subsection{Video Frame Interpolation}
Video Frame Interpolation (VFI) aims at creating non-existent intermediate frames between consecutive frames. 
Many classical methods\cite{4162523,6257455} generally utilize motion estimation and
motion compensation to enhance the temporal resolution of
video sequences. 
In recent years,
deep learning based approaches have achieved
immense success and show their superiority in VFI tasks.
Mainstream learning-based VFI methods can be categorized into two types:  flow-based methods and kernel-based methods.\\ 
\noindent\textbf{ Flow-based methods.} Liu \emph{et al.} \cite{2017Video2} propose a fully convolutional network to estimate voxel flow and hallucinate the intermediate frames. Jiang \emph{et al.} introduce a similar method \cite{0Super} and use
two U-Net architectures to compute bi-directional optical flow and visibility maps to guide the generation of intermediate frames. Bao \emph{et al.} \cite{Wenbo2019MEMC} first linearly weigh the optical flow and then uses a depth-aware projection layer to blend the warped frames adaptively. To estimate large movement, Park \emph{et al.} \cite{park2020robust} take into
account an exceptional motion map that contains the location and
intensity of large motion to approximate a more reliable intermediate frame. Cheng \emph{et al.} \cite{8823006} propose a position feature
transform (PFT) network to estimate the scaling factors that help evaluate different
degrees of the importance of deep features around a target pixel. Different from those backward warping methods, Niklaus \emph{et al.} propose SoftSplat \cite{2020Softmax} to forwardly warp frames and their corresponding feature map using softmax splatting. \\
\noindent \textbf{Kernel-based methods.} Apart from using optical flow, another
major trend for VFI is to replace the two-step interpolation
operation with a convolution process. Niklaus \emph{et al.}\cite{2017Video} use
a pair of 1D kernels to perform a spatially adaptive convolution
to estimate the motion. Cheng \emph{et al.} \cite{2020Video} propose DSepConv
that use deformable separable convolution to enlarge the receptive field of kernel-based methods and further propose EDSC\cite{DBLP:journals/corr/abs-2006-08070} to perform multiple interpolation. Choi \emph{et al.} \cite{2020Channel} propose
CAIN, which implicitly replaces the optical flow computation
module with a channel attention module to capture motion
information.

\subsection{Video Super-Resolution}
Video Super-Resolution (VSR) aims to reconstruct high-resolution videos from the corresponding low-resolution ones. The key to VSR is how to use
complementary information across misaligned frames. The most recent methods can be divided into two categories
according to  generator networks: iterative methods \cite{2018Deep,2019EDVR}
and recurrent methods \cite{2019Recurrent,2020VideoRSDN}. Specifically, TDAN \cite{2020TDAN} adopts deformable convolutions module (DCNs)
\cite{2017Deformable} \cite{zhu2018deformable} to align different frames at the feature level.
EDVR \cite{2019EDVR}, as a representative method, further uses DCNs in a multi-scale pyramid and utilizes multiple attention layers to adopt alignment and then integrate the features. 
PFNL\cite{2020Progressive} utilizes an improved non-local operation to avoid the complex motion estimation and motion compensation, which has obtained a
favorable result in terms of both performance and complexity. Isobe \emph{et al.} \cite{2020VideoTGA}  propose TGA that divides temporal information into groups and utilizes 2D and 3D residual blocks for inter-group fusion from a temporally sliding window. Zhang \emph{et al.} \cite{9292987} propose an efficient 3D
convolutional block (E3DB) for VSR, which  reduces the
computing cost compared with conventional 3D convolution while maximally maintaining the temporal
information.
For the iterative-based method, Tao \emph{el al}. \cite{2015Convolutional} propose a sub-pixel motion compensation layer in a CNN framework and utilize convolutional long-short-term memory (ConvLSTM)  \cite{2015Convolutional} module for capturing temporal information among multiple frames. MTUDB\cite{2019Multi} embeds ConvLSTM structure into ultra-dense residual blocks to construct a multi-temporal ultra-dense memory network for video super-resolution.

Still, most previous iterative-based or temporal sliding window based methods ignore the useful information from subsequent LR frames.  More recently, much work \cite{yi2021omniscient,2020BasicVSR} shows that it is essential to utilize both neighboring LR frames and long-distance LR frames (previous and subsequent) to reconstruct HR frames.

\subsection{Space-Time Video Super-Resolution}
Space-Time Video Super-Resolution (ST-VSR) aims to transform a low spatial resolution video with a low frame rate into a higher spatial and temporal resolution. Some of the earlier methods \cite{2002,2011Space}, which are not based on deep learning, are slow in processing speed and often fail to generate promising effects when processing complex scenarios. Some deep learning-based work has recently made significant progress in speed and effect. STARnet\cite{2020Space} leverages mutually informative relationships between time and space with an optical flow estimation module\cite{2016FlowNet} and performs feature warping of two consecutive frames to interpolate the intermediate frame. Zooming Slow-Mo \cite{2021Zooming} creates a unified framework with  ConvLSTM to align and aggregate temporal information and synthesize the intermediate features via deformable convolution before performing feature fusion for ST-VSR. Based on Zooming Slow-Mo, Xu \emph{et al.} \cite{xu2021temporal} propose a temporal modulation network via a locally-temporal feature comparison module and deformable convolution kernels for controllable feature interpolation, which can interpolate arbitrary intermediate frames.  

However, these ConvLSTM-based solutions are not capable of handling long input sequences. In contrast to simple RNN, LSTM requires to save more intermediate states (hidden state and cell state) in the recurrent process and has a higher computational complexity. A vanilla RNN, on the other hand, only needs to save hidden states. As a result, LSTM often leads to more memory footprints.  Some recent work of VSR\cite{2020BasicVSR,yi2021omniscient} points out that in the VSR task, a longer input video sequence benefits from more long-term information and therefore achieves better performance. With this assumption, the ability of these LSTM-based methods to handle long sequences is severely limited, resulting in a degradation in restoration performance.

\begin{figure*}[t]
	\centering
	\includegraphics[scale=0.540]{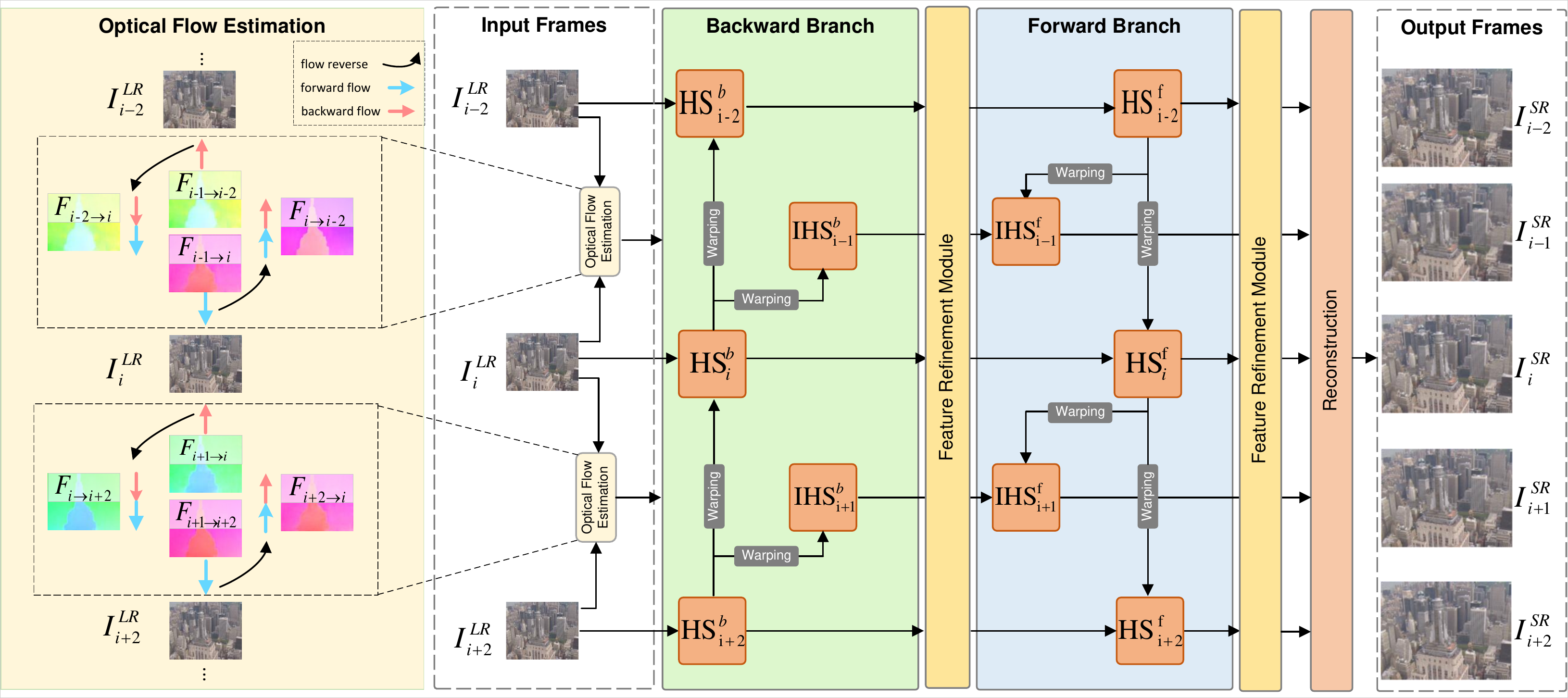}
	\caption{A schematic representation of our proposed method. We first calculate the bidirectional optical flow from the intermediate frames and then reorganize them by optical-flow-reuse strategy. After that, we send LR into a bidirectional recurrent structure, including a forward and backward branch. Afterward, the result of each branch is sent into the Feature Refinement Module (FRM) for further optimization. Lastly, PixelShuffle\cite{2016Real} is applied to reconstruct the SR frames. Note that the feature refinement module is only performed for states of pre-existing frames. } 
	\label{fig:2} 
\end{figure*}

\section{Proposed Method}
In this part, we first demonstrate the optical-flow-reuse strategy, which enables us to simultaneously obtain the motion representation between pre-existing and temporally interpolated frames. Next, we describe how to align features in a bidirectional RNN-style structure. In order to further reduce inaccurate feature alignment and enhance restoration performance, a feature refinement module is then introduced.  In Fig.~\ref{fig:2}, we provide a summary of the proposed approach. An optical flow estimation module is first employed to generate flow that will be used in the bidirectional recurrent network. 
Following that, the candidate LR frames are then routed through a dual branch network, which maintains a continuously updated hidden state. In this process, information obtained from other frames is transmitted via the continuously updated state. After that, a Feature Refinement Module (FRM) is utilized to optimize the hidden states and suppress noise. Finally, an upsampling module is applied to obtain high-resolution output frames. 
\subsection{Optical Flow Estimation Module}\label{flow_method}
As noted in much previous work\cite{2020BasicVSR,yi2021omniscient,DBLP:journals/corr/abs-2006-08070}, the alignment is critical for VSR and VFI. In this section, we first briefly review the alignment process of some representative ST-VSR algorithms before diving into our method.  Zooming SlowMo\cite{2021Zooming} aligns and aggregates temporal information using an end-to-end deformable ConvLSTM framework. It estimates the deformable offset via the two adjacent pre-existing frames and uses the estimated offset to synthesize the intermediate frame. The offset between the pre-existing frames must also be estimated accordingly. TMnet\cite{xu2021temporal} builds a temporal modulation network based on Zooming SlowMo by comparing locally-temporal features to achieve time-arbitrary interpolation. Its pre-existing frame alignment procedure is similar to Zooming SlowMo. When performing time-arbitrary interpolation, each desired intermediate moment also needs to estimate its corresponding offsets.  To summarize, all these methods perform alignment for two kinds of frames separately.

\begin{figure*}[t]
	\centering
	\includegraphics[width=18cm]{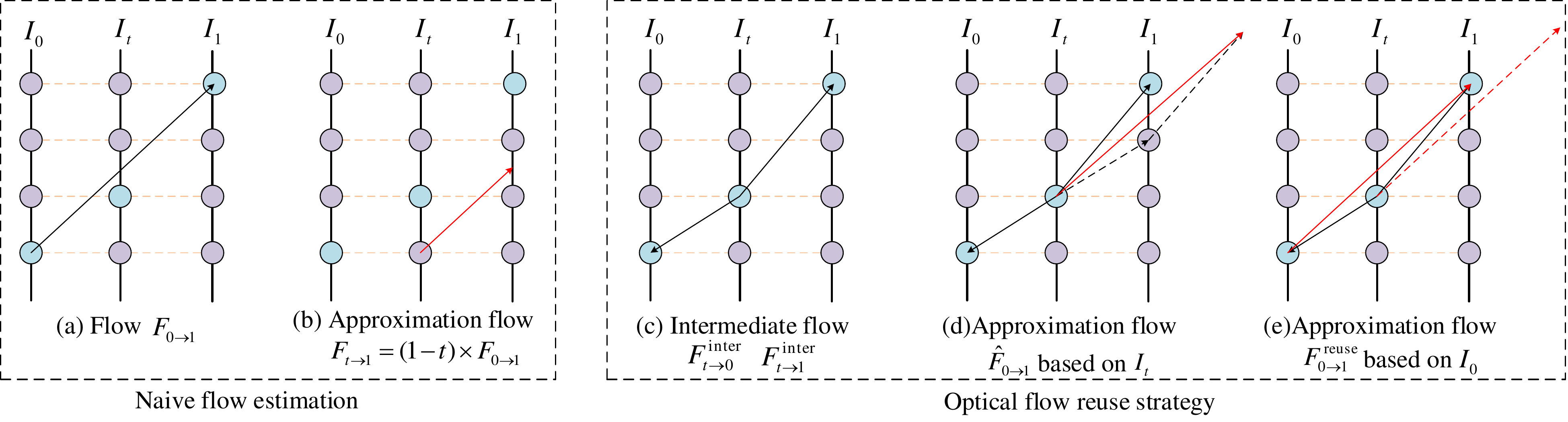}
	\caption{ Comparison of two flow estimation methods: Each vertical line represents a frame, dots here correspond to pixels in the frame. $I_{0}$ and $I_{1}$ are input LR, and $I_{t}$ is an inaccessible intermediate frame. Purple dots denote still background pixels, while blue dots depict moving pixels. For naive flow estimation, the flow $F_{t \rightarrow 1}$ in (b) is obtained using linear multiplication with the flow $F_{0 \rightarrow 1}$ in (a). For intermediate flow estimation, we first calculate the intermediate flow in (c), then, $\hat{F}_{0 \rightarrow 1}$ in (d) is approximated according to the parallelogram rule of vectors. Finally, we employ approximated $F^{com}_{0 \rightarrow t}$ to backward warp the estimated $\hat{F}_{0 \rightarrow 1}$ so that its base position is closer to  $I_{0}$ under the linear motion assumption.    } \label{fig:3}
\end{figure*}

However, it is obvious that the motion information of the pre-existing frames and the temporally interpolated frames are highly correlated, which inspired us to use optical flow to perform alignment more efficiently.  We use $I_{0,1}$ to represent two adjacent frames, and we need to estimate the optical flow between the two 
adjacent frames $F_{0\rightarrow 1}$ and $F_{1\rightarrow 0}$, as well as the  optical flow from the intermediate moment $t$, $F_{t\rightarrow 0}$ and $F_{t\rightarrow 1}$.
Theoretically, we can use any off-the-shelf optical flow estimation method to estimate the optical flow between pre-existing frames, but how to calculate the optical flow towards the intermediate frames has become an intractable problem. Some previous work\cite{2019Quadratic,2020Enhanced} first computes the intermediate optical flow by multiplying flow between pre-existing frames with a hyperparameter  $t \in (0,1)$. Then, a relatively complex post-processing process is introduced to optimize these intermediate flow further. Another option is to estimate the optical flow of the intermediate frames directly, but in this way, we also have to estimate the optical flow between the existing frames.  In the following, we will discuss the two estimation methods and compare their differences. To more clearly explain the proposed optical-flow-reuse strategy, we begin by estimating the optical flow of two adjacent frames. We illustrate the difference between the two flow estimation methods in Fig.~\ref{fig:3}. For the naive flow estimation, we first estimate the optical flow between two adjacent frames with an off-the-shelf method\cite{2017PWC}. Then, we compute the intermediate flow at the moment $t$ by
scaling the magnitude accordingly ($t$ or $1-t$ in (Eq.~\ref{f2})). This process can be described as:
\begin{equation}
	F_{0\rightarrow1} = \phi (I_{0},I_{1}),
	F_{1\rightarrow0} = \phi (I_{1},I_{0}) \label{f1}.
\end{equation}
\begin{equation}
	F_{t\rightarrow0} = t \times \phi (I_{1},I_{0}),
	F_{t\rightarrow1} = (1-t) \times  \phi (I_{0},I_{1}) \label{f2}.
\end{equation}
where $I_{0,1}$ denotes two consecutive frames, and $\phi$ represents an off-the-shelf optical flow estimation model.
It is important to emphasize that this approximation is not so accurate. In addition to the fact that the simple linear estimation cannot accurately fit the nonlinear motion, the starting position of the optical flow estimated in this way is also biased since the flow origin based on $I_{0}$ and $I_{t}$ are actually not at the same location (Fig.~\ref{fig:3} (b)). Much previous  work\cite{2020BMBC,sim2021xvfi,park2021asymmetric} first estimates the intermediate optical flow in this way roughly and then employs a very complex post-processing network to mitigate the optical flow estimation errors. Although these methods achieve 
competitive performance, they also significantly increase  computational complexity. It is impractical to perform such complex processing for every temporally interpolated frame in a recurrent structure.

Unlike the naive flow estimation, we first estimate optical flow based on intermediate time $t$ and then reuse them to approximate the optical flow between the pre-existing input frames. We  use IFnet\cite{huang2020rife} as the optical flow estimation module owning its structural simplicity and computational efficiency.  Unlike some other intermediate optical flow estimators\cite{2020BMBC,park2021asymmetric} that perform dense matching with high computational complexity, IFnet consists of only simple convolution operations. This process can be formulated as:
\begin{equation}
	F^{inter}_{t\rightarrow0},F^{inter}_{t\rightarrow1} =  \Theta (I_{0},I_{1}), \label{f3}
\end{equation}
where $\Theta$ represents the intermediate flow estimator and $F^{inter}_{t \rightarrow \{0,1\}}$ denotes the intermediate flow. Regarding the flow between LR inputs, We first calculate the direction and magnitude of the target optical flow according to the parallelogram law of vectors: 
\begin{equation}
	\hat{F}_{0\rightarrow1} =  - F^{inter}_{t\rightarrow0} + F^{inter}_{t\rightarrow1}, 
\end{equation}
However, as shown in Fig.~\ref{fig:3} (d), the estimated flow $\hat{F}_{0\rightarrow1}$ is currently based on $I_{t}$ rather than $I_{0}$. To alleviate this problem, we introduce a complementary flow $F^{com}_{0\rightarrow t}$ to backward warp $\hat{F}_{0\rightarrow1}$ so that its base position is closer to $I_{0}$. Under the locally smooth assumption of optical flow \cite{2020BMBC}, we have 
$F^{com}_{0\rightarrow t} \approx \frac{t}{1-t}F^{inter}_{t \rightarrow 1}$. As shown in Fig.~\ref{fig:3} (e), we use the estimated complementary flow $F^{com}_{0\rightarrow t}$ to backward warp $\hat{F}_{0\rightarrow1}$  and finally get the approximated $F^{reuse}_{0\rightarrow1}$. This operation can be described as:
\begin{equation}
	F^{reuse}_{0\rightarrow1} =  \mathcal{W}(\hat{F}_{0\rightarrow1},F^{com}_{0\rightarrow t}),
\end{equation}
where $\mathcal{W}$ is the backward warping operator.
The estimation of $F^{resue}_{1\rightarrow0}$ is performed uniformly but independently.
In practice, we observe that such flow correction operation yields slightly better results. 
Although neither of the two estimation methods is entirely accurate,
we found that directly estimating the optical flow of the intermediate frames achieves better performance than the naive flow estimation. Especially for synthetic intermediate frames, they can better profit from nonlinear estimation advantages, thereby synthesizing more reliable intermediate results. At the same time, since operations such as flow flipping consume almost no time, this strategy can significantly shorten the time for motion estimation and directly generate all motion information for alignment with high efficiency. Detailed performance comparisons are provided in Section \ref{ablation}. 

We then discuss how to estimate bidirectional flow across frames.
Given a sequence  $ I^{LR}  = \{{I_{2l-1}^L}\}_{l=1}^{n}$  that contains $n$ consecutive LR frames, 
the target is to produce the forward and backward optical flow between every two adjacent frames. Therefore, we extract two subsequences from the LR sequence,  $ seq\uppercase\expandafter{\romannumeral1}  = \{{I_{2l-1}^L}\}_{l=1}^{n-1}$  and
$ seq\uppercase\expandafter{\romannumeral2} = \{{I_{2l-1}^L}\}_{l=2}^{n}$, and then estimate the optical flow of the corresponding frames in the two subsequences by the proposed optical-flow-reuse approach $\Psi$ as mentioned above. 
This process can be described as:
\begin{equation}
	\begin{aligned}
		F_{l}^{f,b}  &=  \Psi(seq\uppercase\expandafter{\romannumeral1}_{l},seq\uppercase\expandafter{\romannumeral2}_{l}), \\
		&=  \{{F}_{(t \rightarrow 0)}^{b},{F}_{(t \rightarrow 1)}^{f},{F}_{(0 \rightarrow 1)}^{f},{F}_{(1 \rightarrow 0)}^{b} \}_{l},
	\end{aligned}
	\label{f4}
\end{equation}
where intermediate moment $t \in (0,1) $. $F_{l}^{f,b}$ denotes the estimated optical flow between $I_{l}$ and $I_{l+1}$ frame.
So far, the estimation of optical flow has been accomplished. The generated optical flow is employed as the motion information for alignment in the bidirectional recurrence, which will be introduced in detail in the next section.
\subsection{ Bidirectional Recurrent Network} 

In this section, we first introduce the basic structure of the bidirectional recurrent network. Then, we consider applying alignment from coarse (flow warping) to fine (feature refinement module) across frames of the whole input video sequence. 
We use LR and SILR to represent low-resolution inputs and synthetic intermediate frames for brevity.
\subsubsection{Basic structure of bidirectional recurrent network}
We follow the design of some previous ST-VSR work\cite{2021Zooming,xu2021temporal} and use a bidirectional recurrent network to realize global information propagation.
In the recurrent process, the gained information from other frames is transmitted through the continuously updated state. 
Unlike those methods that estimate the synthetic intermediate frames by introducing additional calculations during the recurrent process, the proposed method can directly utilize the optical flow produced in the optical-flow-reuse strategy to align LR and SILR simultaneously. In the proposed bidirectional settings, both frames and features are propagated in two opposite directions, and LR frames in the input sequence could leverage knowledge from any other 
frames. Moreover, any low-resolution synthetic intermediate frame can also leverage the information from neighboring frames or the hidden state. We refer to the two recurrent processes as the forward branch and backward branch, respectively.
As shown in Fig.~\ref{fig:2}, we denote $f$ as the underlying function modeled by our OFR-BRN and take the example of feeding three consecutive LR frames and outputting five SR frames,
\begin{equation}
	\begin{aligned}
		f & : (I^{LR}_{i-2},I^{LR}_{i},I^{LR}_{i+2}), \\
		\rightarrowtail & (I^{SR}_{i-2},I^{SR}_{i-1},I^{SR}_{i+1},I^{SR}_{i+1},I^{SR}_{i+2}).
	\end{aligned}
	\label{f}
\end{equation}
\subsubsection{Backward branch}
The backward branch starts from the last frame of the input sequence and propagates backward continuously.
The recurrent process of LR can be described as:
\begin{equation}
	\begin{aligned}
		HS_{i}^{b} &= \mathcal{W}(HS_{i+2}^{b},F_{ i \rightarrow i+2}^{f}) , \\
		RHS_{i}^{b} &= FRM(HS_{i}^{b},LR_{i}), 
	\end{aligned}
	\label{f6.2}
\end{equation}
where $HS^{b}_{i+2}$ denotes the hidden state in the backward branch which contains information that passed from the preceding state, $\mathcal{W}$ stands for flow warping and  $RHS_{i}^{b}$ represents the refined feature. That is, flow warping is performed first to renew the continuously updated state. Then, the aligned feature and corresponding LR will be fed into the FRM to optimize the hidden state further, which is introduced  in detail in the Section III-C.  Afterward, the feature of the hidden state will be sent into a stacked fusion residual block to get the result of the backward branch.

Next, the recurrent process of the intermediate state will be introduced in the same manner.
Nevertheless, it should be noted that we cannot access the ground truth of the synthesized intermediate frames. So, in the procedure of motion estimation and motion compensation, the propagating states should be not only aligned in feature space but also in frame space and keep consistent with that of pre-existing frames.
Considering that the intermediate frames are inaccessible, it is necessary to find a proper approach to approximate them. Due to the possible occlusion of objects and camera panning in boundary regions of the image, simply incorporating the frames from backward and forward branches may introduce pixel disarrangement. So we employ two masks to reveal these occlusion areas. The frame interpolation process can be formulated as:
\begin{equation}
	\begin{aligned}
		SILR_{i-1} &= \frac{1}{2} \cdot((1-M_{i-2}+M_{i})\otimes \mathcal{W}(LR_{i},{F}_{i-1 \rightarrow i}^{f})\\
		&+ (1-M_{i}+ M_{i-2}) \otimes \mathcal{W}(LR_{i-2},{F}_{i-1 \rightarrow i-2}^{b}) ),  
	\end{aligned}
	\label{f9}
\end{equation}
where $M_{i-2}$ and $M_{i}$ denotes two masks which reveal these occlusion areas between the two adjacent frames $LR_{i-2}$ and $LR_{i}$. Notice that points with a value of 0 in the mask refer to those pixels that exist in the estimated intermediate frame and disappear due to occlusion or movement in $LR_{i-2}$ and $LR_{i}$,
where $\otimes$ denotes the Hadamard product. We follow the assumptions of SuperSloMo\cite{0Super} that if a pixel $p$ is visible at an intermediate moment, it is most likely
at least visible in one of the two input images. So, in short, this operation means that for either of the two 
warped frames, if a pixel point no longer exists due to occlusion or movement, it is supplemented by the pixel point in the other image at the corresponding position. 
Similar to LR, the backward branch of intermediate hidden state (IHS) can be described as:
\begin{equation}
	\begin{aligned}
		IHS_{i-1}^{b} &= \mathcal{W}(HS_{i}^{b},F_{i-1 \rightarrow i}^{f}), \\
		FIHS_{i-1}^{b} &= Fuse(IHS_{i-1}^{b},SILR_{i-1}),\\  
	\end{aligned}
	\label{f7}
\end{equation}
where $HS_{i}^{b}$ is aligned to $IHS_{i-1}^{b}$ via flow warping and $Fuse$ means fusing the SILR in frame and feature space through  cascading residual blocks to get the fusion result $FIHS_{i-1}^{b}$. Note that we only use a simple fuse operation to blend the features for the temporally interpolated states. Since the FRM  requires both the hidden state and the corresponding LR as input, we only perform feature refinement for frames that originally exist in the input sequence. 
\subsubsection{Forward branch}
The forward branch updates the state from front to back in chronological order.
It roughly shares the same idea with that of backward ones.The recurrent processes of LR can be described as:
\begin{equation}
	\begin{aligned}
		HS_{i}^{f} &= \mathcal{W}(HS_{i-2}^{f},F_{i \rightarrow i-2}^{b}), \  \\
		RHS_{i}^{f} &= FRM(HS_{i}^{f},LR_{i}),\\  
	\end{aligned}
	\label{f6}
\end{equation}
where $HS_{i-2}^{f}$ is aligned to $ HS_{i}^{f}$ by optical flow $F_{i \rightarrow i-2}^{b}$. Then, the aligned feature and corresponding LR are fed into FRM to refine the feature further.
The forward branch of intermediate hidden state (IHS) can be described as:
\begin{equation}
	\begin{aligned}
		IHS_{i-1}^{f} &= \mathcal{W}(HS_{i-2}^{f},F_{i-1 \rightarrow i-2}^{b}), \\
		FIHS_{i-1}^{f} &= Fuse(IHS_{i-1}^{f},SILR_{i-1}),\\  
	\end{aligned}
	\label{f8}
\end{equation}
This process is basically the same as that of Eq.~\ref{f7}.
Afterward, the obtained $RHS^{f}$ (or $FIHS^{f}$) are first concatenated with their backward counterparts $RHS^{b}$ (or $FIHS^{b}$) and then fed into several residual blocks to get the final features that are ready to be super-resolved. 
\subsection{Feature Refinement Module} 
\begin{figure}[t]
	\centering
	\includegraphics[width=8.5cm]{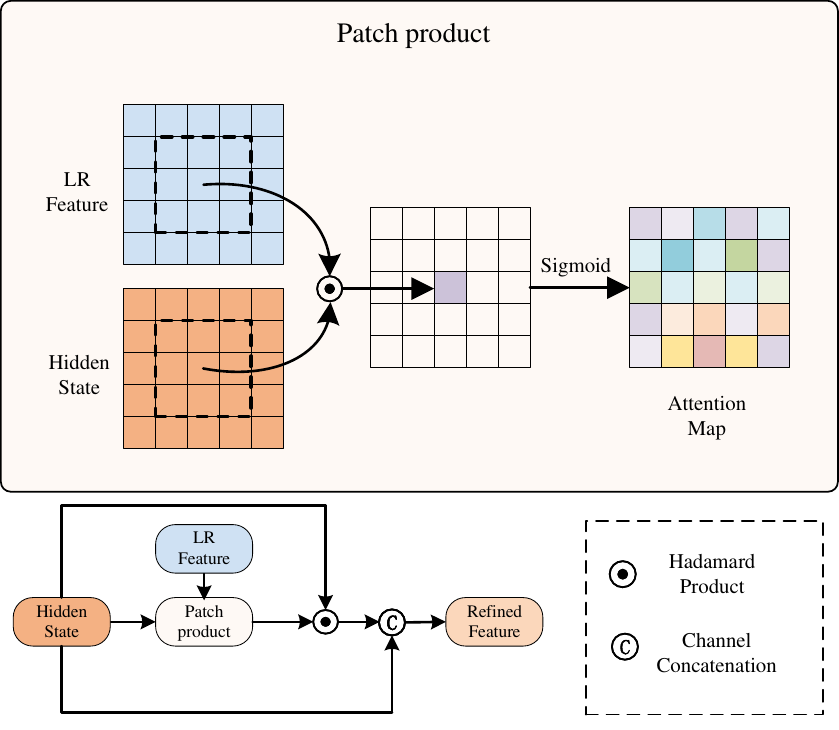}
	\caption{ Illustration of the Feature Refinement Module (FRM). We first compute the local correlation limited in the square patch and then feed the result into a sigmoid function to get the attention map. 
		Next, the hidden state is multiplied by the attention map and a skip-connection structure is employed to facilitate the 
		feature blending. } \label{fig:frm}
\end{figure}
Based on the bidirectional recurrent structure, any frame in the input sequence can obtain information gained from any other frame. However, the continuously updated state introduces not only information gain but also some noise caused by inaccurate alignment and occlusion.  In order to solve this problem, one of the most  straightforward solutions is to replace RNN with LSTM, which can adaptively select the information throughout the input sequence and suppress noise. 
However, because LSTM-based structures might consume more memory, their ability to handle long-range frames is severely limited.  (Detailed comparison results are given in  Section.~\ref{exp}. ) 
Based on the above discussion, we designed a task-oriented module that can be viewed as a lightweight version of LSTM.  It can  measure the relevance between the current candidate frame and the hidden state adaptively to determine which parts should be highlighted and which parts should be suppressed. 
Inspired by\cite{DBLP:journals/corr/BrabandereJTG16,2019Context}, much work has studied how to dynamically generate a convolution kernel and calculate the similarity of two input tensors. Jia \emph{et al.} \cite{2020VideoRSDN} propose RSDN  and design a hidden state adaptation module that allows the current frame to selectively useful information from the hidden state.
Li \emph{et al.} \cite{2020MuCAN} propose MuCAN, which performs a temporal multi-correspondence aggregation strategy and cross-scale nonlocal-correspondence aggregation scheme to explore the self-similarity of images across scales. 
\par
Inspired by \cite{2016FlowNet}, which utilizes a correlation layer that performs  multiplicative patch comparisons between two feature maps, we compute the correlation between the feature map of LR and aligned hidden state, as shown in Fig.~\ref{fig:frm}. Specifically, we feed LR into a “Conv-LeakyReLU” layer and project the feature representations to the same space dimension as the hidden state. Then, we compute the local correlation of the LR feature and hidden state for each channel. The correlation of LR feature ($F_{LR}$)
centered at $x$ and hidden state ($HS$) centered at $y$ in $HS$ map can
be described as:

\begin{equation}
	corr(x,y) =  \sum_{o \in [-k,k]\times[-k,k]}  \left<F_{LR}(x+o),HS(y+o)\right>,
	\label{f10}
\end{equation}
where we set $k$=1 here, and $o$ represents offset that is limited in a square patch. It is worth highlighting that unlike those flow estimation methods\cite{teed2020raft,jia2021braft} which compute all-pair cost volume between two inputs, here, each small patch only calculates the correlation with its counterpart at the same spatial location, which is very time and memory efficient.  After that, we apply the sigmoid
activation function of correlation matrix $corr$  and transform it into a matrix whose value is in the range [0, 1] and then perform element-wise multiplication between the hidden state and matrix to get the optimized hidden feature. Finally, we concatenate the optimized results and the hidden states in the channel dimension. Noticing that as we cannot get the ground truth of SILR, feature refinement is only performed for pre-existing frames. 

Until now, the spatial-temporal features of LR and SILR have been exploited.
Then, we perform spatial upsampling reconstruction for the features. Specifically, the reconstructed feature maps are fed into 
two sub-pixel upscaling modules with PixelShuffle \cite{2016Real} and finally output the reconstructed
HR video frames.

\subsection{ Implementation Details}
We adopt IFNet\cite{huang2020rife} pre-trained  on the Vimeo\cite{DBLP:journals/corr/abs-1711-09078} dataset as our flow estimation module.
For data argument, we randomly crop a sequence of down-sampled image patches with the size of
64$\times$64 and take the odd-indexed LR as inputs, the corresponding consecutive HR
sequence with the size of 256$\times$256 as supervision. During training, we adopt Adam\cite{kingma2014adam} optimizer  with $\beta1$ = 0.9 and $\beta2$ = 0.999
and apply standard augmentation, such as rotation,
flipping, and random cropping. To allow longer propagation, we follow the setup of BasicVSR\cite{2020BasicVSR} and perform temporal augment by flipping the original input sequence. The initial learning rates of the flow estimator and other parts are set to $2.5 \times 10 ^{-5}$
and $1 \times 10^{-4}$ respectively and decay to $1 \times 10 ^{-7}$ with a cosine annealing\cite{2016SGDR}. The batch size is set to be
24, and we train the model on 8 Nvidia 1080Ti GPUs.

\section{Experiments}


\label{sec:bounded-levels}

In this section, we conduct experiments on four
mainly used datasets for VSR and VFI: Vid4\cite{5995614}, SPMCS\cite{tao2017detail}, Vimeo-90k-T\cite{DBLP:journals/corr/abs-1711-09078}, and REDS\cite{Nah_2019_CVPR_Workshops}.
\subsection{Experimental Setup}
\begin{table*}[]
	\centering
	\caption{ Comparison of PSNR (dB), SSIM, speed (FPS), and parameters (million) on  Vid4\cite{5995614}, REDS\cite{Nah_2019_CVPR_Workshops}, SPMCS\cite{tao2017detail}, and Vimeo-90K-T\cite{DBLP:journals/corr/abs-1711-09078}.  \textcolor{red}{Red} and
		\textcolor{blue}{blue} colors indicate the best and the second-best performance. The inference speed is achieved on the Vid4 dataset when performing space $4 \times$ and time $2 \times$ super-resolution using one Nvidia 1080ti GPU.}
	\resizebox{0.97\linewidth}{!}{
		\begin{tabular}{c|cc|cc|ll|cc|cc|cc|cc|c|c}
			\hline
			VFI+(V)SR/ST-VSR methods & \multicolumn{2}{c|}{\begin{tabular}[c]{@{}c@{}}Vid4\\ PSNR SSIM\end{tabular}} & \multicolumn{2}{c|}{\begin{tabular}[c]{@{}c@{}}REDS\\ PSNR SSIM\end{tabular}} & \multicolumn{2}{c|}{\begin{tabular}[c]{@{}c@{}}SPMCS\\ PSNR SSIM\end{tabular}} & \multicolumn{2}{c|}{\begin{tabular}[c]{@{}c@{}}Vimeo-Fast\\ PSNR SSIM\end{tabular}} & \multicolumn{2}{c|}{\begin{tabular}[c]{@{}c@{}}Vimeo-Medium\\ PSNR SSIM\end{tabular}} & \multicolumn{2}{c|}{\begin{tabular}[c]{@{}c@{}}Vimeo-Slow\\ PSNR SSIM\end{tabular}} & \multicolumn{2}{c|}{\begin{tabular}[c]{@{}c@{}}Vimeo-Total\\ PSNR SSIM\end{tabular}} & \begin{tabular}[c]{@{}c@{}}Speed\\ FPS\end{tabular} & \begin{tabular}[c]{@{}c@{}}Parameters\\ millions\end{tabular} \\ \hline
			SuperSloMo\cite{0Super} + Bicubic      & 22.84                                & 0.5772                                 & 25.23                                & 0.6761                                 & 27.02                                 & 0.7456                                & 31.88                                   & 0.8793                                    & 29.94                                    & 0.8477                                     & 28.73                                   & 0.8102                                    & 29.99                                    & 0.8449                                    & -                                                   & 19.8                                                          \\
			SuperSloMo\cite{0Super} + RCAN\cite{2018Image}         & 23.78                                & 0.6397                                 & 26.37                                & 0.7209                                 & 29.63                                 & 0.8325                                & 34.52                                   & 0.9076                                    & 32.50                                    & 0.8844                                     & 30.69                                   & 0.8624                                    & 32.44                                    & 0.8835                                    & 1.91                                                & 19.8+16.0                                                     \\
			SuperSloMo\cite{0Super} + RBPN\cite{2019Recurrent}          & 23.76                                & 0.6362                                 & 26.48                                & 0.7281                                 & 29.15                                 & 0.8488                                & 34.73                                   & 0.9108                                    & 32.79                                    & 0.8930                                     & 30.48                                   & 0.8354                                    & 32.62                                    & 0.8839                                    & 1.55                                                & 19.8+12.7                                                     \\
			SuperSloMo\cite{0Super} + EDVR\cite{2019EDVR}         & 24.40                                & 0.6706                                 & 26.26                                & 0.7222                                 & 29.03                                 & 0.8593                                & 35.05                                   & 0.9136                                    & 33.85                                    & 0.8967                                     & 30.99                                   & 0.8673                                    & 33.45                                    & 0.8933                                    & 4.94                                                & 19.8+20.7                                                     \\ \hline
			Sepconv\cite{2017Video} + Bicubic          & 23.51                                & 0.6273                                 & 25.17                                & 0.6760                                 & 27.10                                 & 0.7494                                & 32.27                                   & 0.8890                                    & 30.61                                    & 0.8633                                     & 29.04                                   & 0.8290                                    & 30.55                                    & 0.8602                                    & -                                                   & 21.7                                                          \\
			Sepconv\cite{2017Video} + RCAN\cite{2018Image}            & 24.92                                & 0.7236                                 & 26.21                                & 0.7177                                 & 29.68                                 & 0.8370                                & 34.97                                   & 0.9195                                    & 33.59                                    & 0.9125                                     & 32.13                                   & 0.8967                                    & 33.50                                    & 0.9103                                    & 1.86                                                & 21.7+16.0                                                     \\
			Sepconv\cite{2017Video} + RBPN\cite{2019Recurrent}            & 26.08                                & 0.7751                                 & 26.32                                & 0.7254                                 & 29.34                                 & 0.8572                                & 35.07                                   & 0.9238                                    & 34.09                                    & 0.9229                                     & 32.77                                   & 0.9090                                    & 33.97                                    & 0.9202                                    & 1.51                                                & 21.7+12.7                                                     \\
			Sepconv\cite{2017Video} + EDVR\cite{2019EDVR}            & 25.93                                & 0.7792                                 & 26.14                                & 0.7205                                 & 29.19                                 & 0.8660                                & 35.23                                   & 0.9252                                    & 34.22                                    & 0.9240                                     & 32.96                                   & 0.9112                                    & 34.12                                    & 0.9215                                    & 4.96                                                & 21.7+20.7                                                     \\ \hline
			DAIN\cite{DAIN} + Bicubic            & 23.55                                & 0.6268                                 & 25.22                                & 0.6783                                 & 27.10                                 & 0.7491                                & 32.41                                   & 0.8910                                    & 30.67                                    & 0.8636                                     & 29.06                                   & 0.8289                                    & 30.61                                    & 0.8607                                    & -                                                   & 24.0                                                            \\
			DAIN\cite{DAIN} + RCAN\cite{2018Image}               & 25.03                                & 0.7261                                 & 26.33                                & 0.7233                                 & 29.68                                 & 0.8370                                & 35.27                                   & 0.9242                                    & 33.82                                    & 0.9146                                     & 32.26                                   & 0.8974                                    & 33.73                                    & 0.9126                                    & 1.84                                                & 24.0+16.0                                                     \\
			DAIN\cite{DAIN} + RBPN\cite{2019Recurrent}               & 25.96                                & 0.7784                                 & 26.57                                & 0.7344                                 & 29.44                                 & 0.8577                                & 35.55                                   & 0.9300                                    & 34.45                                    & 0.9262                                     & 32.92                                   & 0.9097                                    & 34.31                                    & 0.9234                                    & 1.43                                                & 24.0+12.7                                                     \\
			DAIN\cite{DAIN} + EDVR\cite{2019EDVR}               & 26.12                                & 0.7836                                 & 26.39                                & 0.7291                                 & 29.23                                 & 0.8663                                & 35.81                                   & 0.9323                                    & 34.66                                    & 0.9281                                     & 33.11                                   & 0.9119                                    & 34.52                                    & 0.9254                                    & 4.00                                                   & 24.0+20.7                                                     \\ \hline
			RIFE\cite{huang2020rife} + BasicVSR\cite{2020BasicVSR}            & \multicolumn{1}{l}{26.00}            & \multicolumn{1}{l|}{0.7891}            & \multicolumn{1}{l}{26.71}            & \multicolumn{1}{l|}{0.7413}            & 30.07                                 & 0.8607                                & \multicolumn{1}{l}{35.26}               & \multicolumn{1}{l|}{0.9281}               & \multicolumn{1}{l}{33.78}                & \multicolumn{1}{l|}{0.9198}                & \multicolumn{1}{l}{32.22}               & \multicolumn{1}{l|}{0.8993}               & \multicolumn{1}{l}{33.69}                & \multicolumn{1}{l|}{0.9169}               & \multicolumn{1}{l|}{\textcolor{blue}{21.88}}        & 20.9+6.3                                                      \\ \hline
			STARnet\cite{2020Space}                  & 26.06                                & \textcolor{blue}{0.8046}                                 & 26.39                                & 0.7444                                 & 29.24                                 & 0.8478                                & 36.19                                   & 0.9368                                    & 34.86                                    & 0.9356                                     & 33.10                                   & \textcolor{blue}{0.9164}                  & 34.71                                    & 0.9318                                    & 10.54                                               & 111.61                                                        \\
			Zooming Slow-Mo\cite{2021Zooming}          & 26.31                                & 0.7976                                 & 26.72                                & 0.7453                                 & 30.15                                 & 0.8731                                & 36.81                                   & 0.9415                                    & 35.41                                    & 0.9361                                     & 33.36                                   & 0.9138                                    & 35.21                                    & 0.9323                                    & 12.40                                                & \textcolor{red}{11.1}                                         \\
			TMnet\cite{xu2021temporal}                    & \textcolor{blue}{26.43}              & 0.8016               & \textcolor{blue}{26.81}              & \textcolor{blue}{0.7476}               & \textcolor{blue}{30.37}               & \textcolor{blue}{0.8757}              & \textcolor{blue}{37.04}                 & \textcolor{blue}{0.9435}                  & \textcolor{blue}{35.60}                  & \textcolor{blue}{0.9380}                   & \textcolor{blue}{33.51}                 & 0.9159                                    & \textcolor{blue}{35.39}                   & \textcolor{blue}{0.9343}                   & 11.60                                                & 12.26                                                         \\ \hline
			OFR-BRN (Ours)           & \textcolor{red}{26.72}               & \textcolor{red}{0.8141}                & \textcolor{red}{27.30}               & \textcolor{red}{0.7568}                & \textcolor{red}{30.77}                & \textcolor{red}{0.8765}               & \textcolor{red}{37.32}                  & \textcolor{red}{0.9465}                   & \textcolor{red}{35.72}                   & \textcolor{red}{0.9393}                    & \textcolor{red}{33.58}                  & \textcolor{red}{0.9167}                   & \textcolor{red}{35.53}                   & \textcolor{red}{0.9358}                   & \textcolor{red}{21.95}                              & \textcolor{blue}{11.77}                                       \\ \hline
		\end{tabular}
	}
	\label{tab:2}
\end{table*}

\begin{figure*}[t]
	\centering
		\includegraphics[width=16cm]{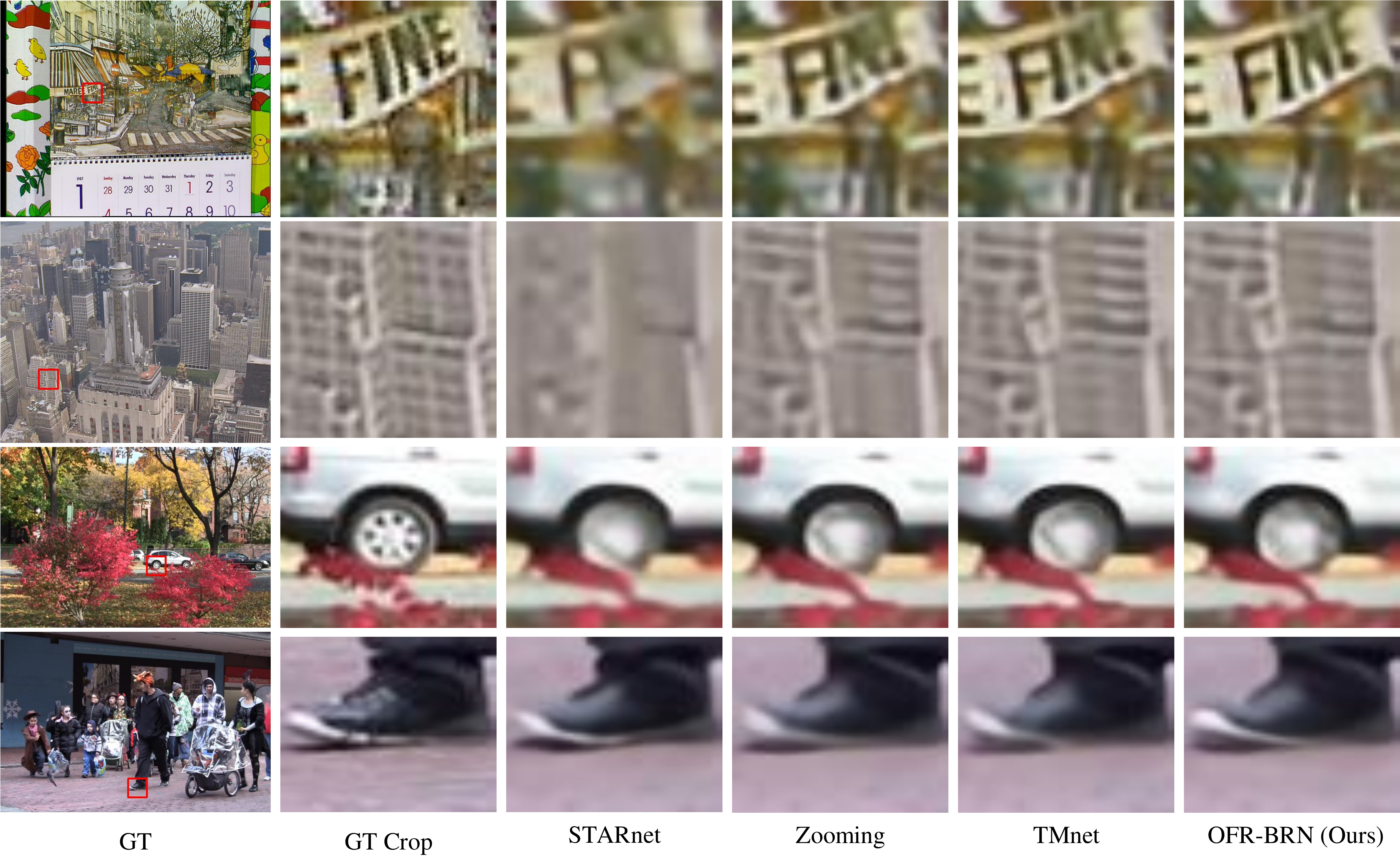}
	\caption{ Qualitative comparison on Vid4\cite{5995614}. 
		Part of the areas are zoomed in and framed with a \textcolor{red}{red} box to facilitate comparison.
	}\label{vid}
	\centering
\end{figure*}
\begin{table*}[t]
	
	\caption{ Comparison of  performance (PSNR(dB))$\uparrow$ / average memory usage (MB)$\downarrow$  on REDS when inputting different lengths of LR frames. The red color indicates the best performance. Note that we perform space 4$\times$ and time 2$\times$ interpolation for the input LR frames at the resolution of 180$\times$320 using one Nvidia 1080Ti GPU. N/A corresponds to out-of-memory cases. }
	\resizebox{1.00\linewidth}{!}{\begin{tabular}{c|c|c|c|c|c|c|c|c|c}
			\hline
			\begin{tabular}[c]{@{}c@{}}Input\\ length\end{tabular} & 4            & 5            & 6            & 7             & 10           & 16           & 22           & 25           & 51           \\ \hline
			TMNet\cite{xu2021temporal}                                                  & 26.76 / 7590 & 26.82 / 7830 & 26.81 / 9692 & 26.81 / 10904 & N/A            & N/A            & N/A            & N/A            & N/A            \\ \hline
			Zooming
			Slow-Mo\cite{2021Zooming}& 26.68 / 6096  & 26.72 / 7417 & 26.73 / 9154 &  26.72 / 10590 & N/A            & N/A            & N/A            & N/A            & N/A            \\ \hline
			Ours                                                   & \textcolor{red}{27.07} / \textcolor{red}{2226} & \textcolor{red}{27.09} / \textcolor{red}{2312} &  \textcolor{red}{27.10} / \textcolor{red}{2468} & \textcolor{red}{27.11} / \textcolor{red}{2568} & \textcolor{red}{27.13} / \textcolor{red}{2648} & \textcolor{red}{27.15}  / \textcolor{red}{3780} &\textcolor{red}{27.17} / \textcolor{red}{4468} & \textcolor{red}{27.18} / \textcolor{red}{4902} & \textcolor{red}{27.30} / \textcolor{red}{8802} \\ \hline
	\end{tabular}}
	\label{tab:length}
\end{table*}
\noindent \textbf{Datasets}
We adopt the Vimeo-90K-T septuplet trainset \cite{DBLP:journals/corr/abs-1711-09078} for training. 
Vimeo-90K-T contains 91,701 video sequences, each of which
consists of 7 frames, and the HR frames are at the 
resolution of 448 $\times$ 256. We follow the setting of Zooming SloMo\cite{2021Zooming} and divide the testing dataset of  Vimeo-90K-T into three categories according to the average motion
flow magnitude: fast motion, medium motion, and slow motion, which include 1225, 4977, and 1613 video clips, respectively.
For a fair comparison, we remove  5 video clips
from the original medium-motion set and 3 clips from the slow-motion set because these sets contain only all-black backgrounds,
which will lead to infinite values on PSNR. We also test on Vid4, which contains
four scenes, and this dataset is widely used in VSR tasks. Finally, to verify the robustness of our method across different datasets, we  test the model on REDS and SPMCS. These test sets are more challenging because they contain large motions and diverse scenarios.
To train our model, we generate LR frames with a bicubic downsampling factor of 4 and use odd-indexed LR frames as input to predict the
corresponding consecutive HR outputs.\\
\textbf{Loss function}
To optimize our model, we follow Zooming SlowMo\cite{2021Zooming} and use a Charbonnier penalty function\cite{lai2017deep} as the reconstruction loss function:
\begin{equation}
	\begin{aligned}
		loss_{rec} = \sqrt{{\Vert I^{res}-I^{GT} \Vert}^{2}+{\epsilon}^{2}},
	\end{aligned}
	\label{loss}
\end{equation}
where $I^{res}$ refers to the restoration outputs and $I^{GT}$ denotes ground-truth HR video frames. $\epsilon$ is a  constant value, and we empirically set it to 1 $\times$ 10$^{-3}$.\\
\textbf{Evaluation}
We adopt Peak Signal-to-Noise Ratio (PSNR) and Structural Similarity Index\cite{wang2004image} (SSIM) for evaluation 
which  are widely used  for VFI and VSR. We also provide model size (Million) and inference FPS of different methods to compare efficiency.
Note that the inference FPS is computed on the entire Vid4  dataset measured with one Nvidia 1080Ti GPU.
\subsection{ Comparison with State-of-the-art methods}\label{exp}

\noindent \textbf{Comparison methods}
We compare the performance of our model with
some state-of-the-art two-stage methods (VFI and VSR) and one-stage ST-VSR methods.
For the two-stage  methods, we perform video frame interpolation (VFI) by SuperSloMo\cite{0Super}, DAIN\cite{DAIN} 
SepConv\cite{2017Video} or RIFE\cite{huang2020rife} and perform super-resolution by Bicubic Interpolation (BI), RCAN\cite{2018Image}, RBPN\cite{2019Recurrent}, EDVR\cite{2019EDVR} or BasicVSR\cite{2020BasicVSR}. For one-stage ST-VSR models, we compare our network with recently state-of-the-art methods
Zooming Slow-Mo\cite{2021Zooming}, STARnet\cite{2020Space} or TMnet\cite{xu2021temporal}. When training, we use Vimeo-90K-T trainset\cite{DBLP:journals/corr/abs-1711-09078} and feed odd LR frames into the model and reconstruct HR frames corresponding to the frames of the entire sequence.
All these methods are trained on the Vimeo-90K training set and evaluated on the
Vimeo-90K test set, Vid4\cite{5995614}, REDS\cite{Nah_2019_CVPR_Workshops} test set and SPMCS\cite{tao2017detail}. \\
\noindent \textbf{Comparison on Vimeo-90k-T and Vid4. } We first analyze the restoration results on Vimeo-90k and Vid4, which contain frames with relative small motion magnitudes.
The quantitative results are listed in Table~\ref{tab:2}, and the visual comparisons are given in Fig.~\ref{vid}. 
Following the suggestion of Zooming SlowMo\cite{2021Zooming}, we omit baseline models with bicubic interpolation when comparing
the speed. One can see that it outperforms the second-best method by 0.29dB on Vid4 and runs 2$\times$ faster than other state-of-the-art ST-VSR methods.
On the Vimeo dataset, the proposed model also outperforms other methods
on all qualitative evaluation indicators.
It can be seen that the advantages of the proposed method are not apparent when the motion magnitude is small, and the image texture is relatively simple (Vimeo-Slow). However, the proposed model has more advantages when reconstructing sequences with larger motion magnitudes (Vimeo-Fast) and more complex texture details (Vid4).
In addition, our model size remains close to other state-of-the-art  methods.  \\
\noindent \textbf{Comparison on REDS and SPMCS.}
In order to further verify the robustness of our method across different datasets, we trained the model on Vimeo and tested it on REDS and SPMCS.   REDS and SPMCS have a higher resolution, more complex and diverse scenes, and more significant motion than Vimeo and Vid4.
Specifically, we set up the test setting with reference to \cite{2019EDVR}, and compared the
PSNR, SSIM  of different schemes.
\begin{figure*}[htbp]
	\centering
		\includegraphics[width=16cm]{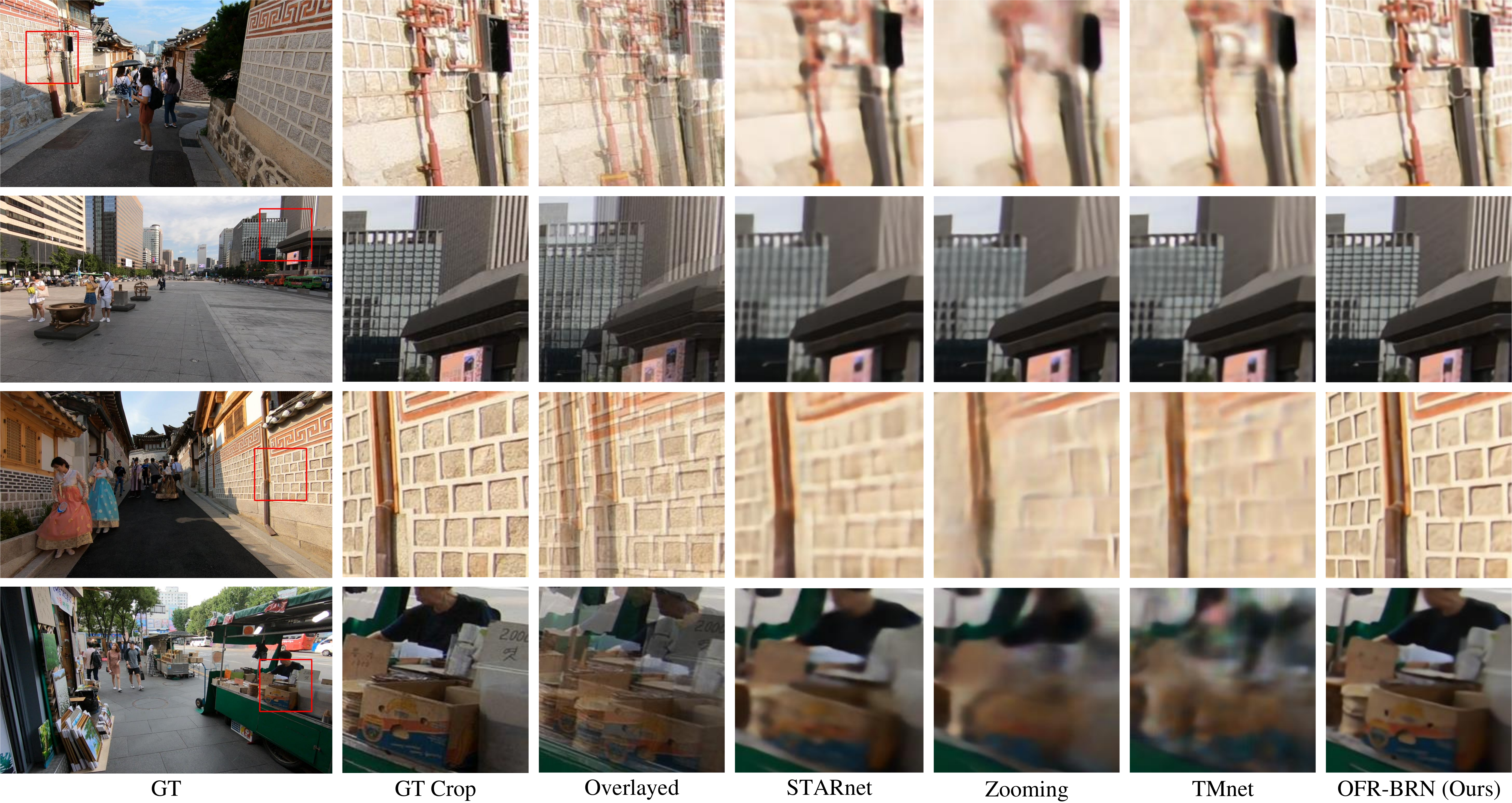}
	\caption{  Qualitative comparison of the synthetic intermediate frames on REDS\cite{Nah_2019_CVPR_Workshops}.
		The overlayed crop column shows the two overlapping frames of the adjacent inputs. Part of the areas are zoomed in and framed with a \textcolor{red}{red} box to facilitate comparison.
	}\label{fig:reds}
	
	\centering
	
\end{figure*}
It can be seen in Table \ref{tab:2} that OFR-BRN outperforms other ST-VSR methods in accuracy and speed. When comparing with the two-stage strong baseline RIFE\cite{huang2020rife}+BasicVSR\cite{2020BasicVSR}, our approach shows great advantage. Notice that RIFE employs the same flow estimation structure (IFnet) as ours.  It first uses IFnet (about 6.8M) to estimate the intermediate optical flow of two adjacent frames. Then it introduces a complex post-processing module (about 14.1M) to synthesize the intermediate frame.   In contrast, our model does not adopt complex post-processing modules after flow warping except for two masks. We assume that the performance improvement is mainly due to our proper usage of long-range temporal information when synthesizing the intermediate frames.
Additionally, the proposed method can often achieve more visually appealing results when synthesizing intermediate frames. We provide more visual comparisons of synthetic intermediate frames on the REDS\cite{Nah_2019_CVPR_Workshops} dataset in Fig.~\ref{fig:reds}.  Since some scenes in the REDS contain huge motions and severe camera panning, these scenes can better demonstrate the model's ability to handle extreme movements. It can be seen that there exist huge motions between two adjacent frames, some kernel-based methods (Zooming Slow-Mo, TMNet) appear to have obvious blurring, and some grid-like parts are severely distorted.
In contrast, the proposed method can approximate more compelling qualitative results by benefiting from the intermediate flow estimation tactics. 
This also shows that estimating the optical flow from the intermediate frame can better deal with the non-linear motion to obtain better results.\\
\noindent \textbf{Comparision of long-term modeling ability.}
Much previous VSR\cite{2020BasicVSR,yi2021omniscient} and VFI\cite{kalluri2021flavr} work has established the fact that the proper use of multi-frame information can improve the restoration performance. However, we cannot input LR frames of arbitrary length at a time due to GPU memory constraints. Therefore, from a practical point of view, it is essential to compare the performance and efficiency when inputting different lengths of LR frames. 
We list the comparison results of performance (PSNR(dB)) / average memory usage (MB)   on REDS\cite{Nah_2019_CVPR_Workshops} dataset of different methods in Table~\ref{tab:length}. The comparison results can be summarized in three points. 1) Our approach achieves higher performance at different input lengths while consuming less GPU memory. 2) Our method is capable of dealing with very long input sequences. In contrast, Zooming Slow-Mo\cite{2021Zooming}, TMNet\cite{xu2021temporal} excess the GPU memory just at the length of 8. 3) The performance of our method increases when the input length increases, which indicates that our model can better utilize the information within long-range frames. Still, the performance of TMNet or Zooming Slow-Mo does not show a strict positive correlation with the input length. \\
\begin{figure}[t]
	\centering
		\includegraphics[width=9cm]{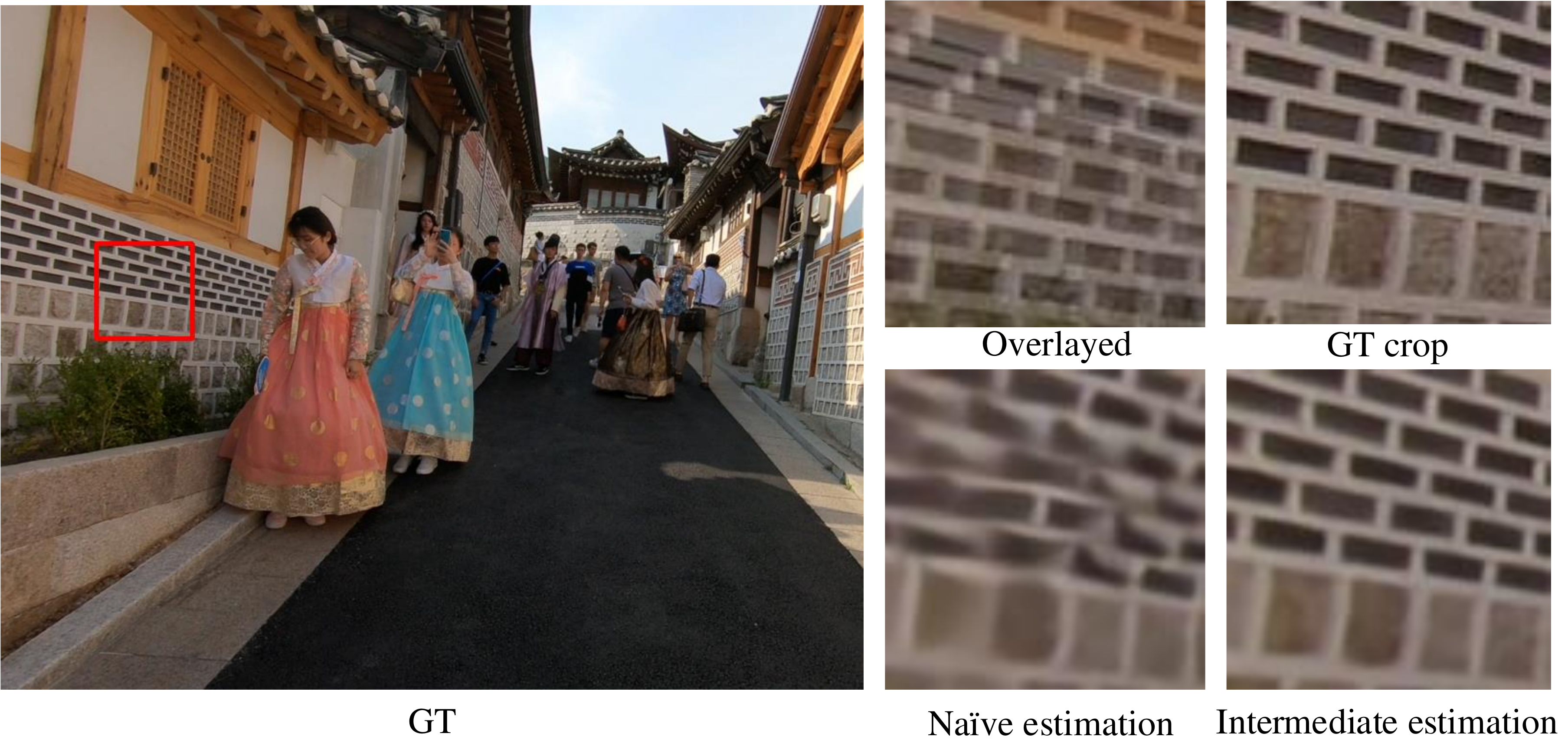}
	\caption{ Visual comparison of intermediate flow estimation and naive flow estimation. 
		The overlayed image shows the overlapping frames of the two adjacent inputs.
	}\label{fig:5}
	\centering
	\vspace{-1em}
\end{figure}

\noindent \textbf{More discussion about memory usage and performance.}
It is necessary to explain why the proposed method is faster and more memory-saving when handling long input sequences. We attribute this mainly to the following factors:
First, the optical-flow-reuse strategy allows the proposed model to simultaneously estimate the motion information between the LR inputs and the intermediate frames. However, most previous ST-VSR approaches estimate both motion information separately, which does not exploit the relationship between VFI and VSR well. In terms of network structure, we followed the bidirectional recurrent structure from \cite{2021Zooming,xu2021temporal}. Still, we discarded the ConvLSTM\cite{2015Convolutional} design because this structure needs to maintain more intermediate states during inference. Instead, we use a more lightweight module (FRM) for feature optimization. 
FRM itself does not have any learnable parameters except for several convolution layers for feature blending and does not need to save any intermediate states. Finally, the frame interpolation module in our model is straightforward compared to the various complex frame interpolation designs. We opt not to design complex optical flow refinement operations\cite{huang2020rife} or context refilling modules\cite{2018Context,2020Softmax} to ensure a simple and efficient network that can handle longer input sequences.
\subsection{Ablation Study}\label{ablation}
\begin{figure}[t]
	\centering
	\includegraphics[width=8.5cm]{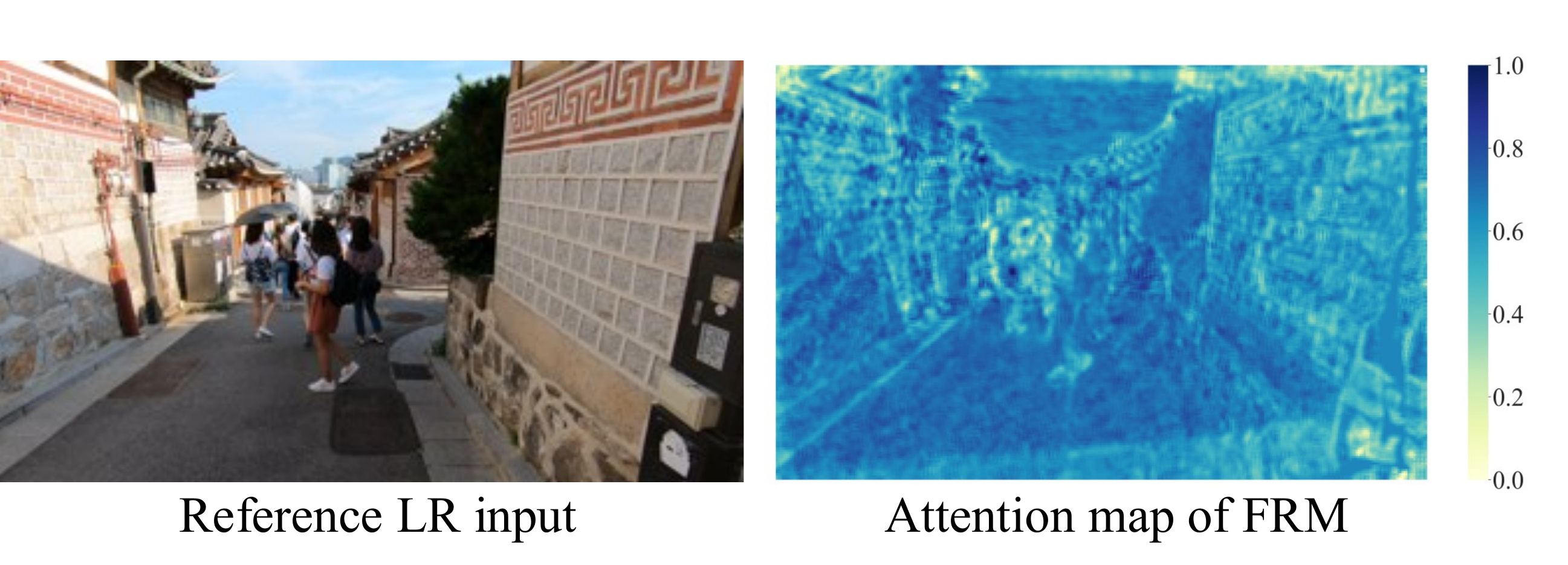}
	\caption{ Visualization of FRM's attention map. Regions with higher similarity tend to have more attention, indicating that the corresponding areas in the hidden state are more informative. Note that we first normalize the values of the attention map for better visualization. } \label{fig:frm_vis}
\end{figure}

After comparing with the existing work, we conduct examinations of different modules of
the proposed structure.
Specifically, we mainly focus on 1) different strategies of recurrence
2) different methods of flow estimation
3) the effectiveness of Feature Refinement Module (FRM)  4) influences of feature fusion space.
The results of the ablation study on different modules are listed in Table \ref{tab:5}. \\
\textbf{ One-way recurrence vs. bidirectional recurrence.}
To test the usefulness of the bidirectional mechanism, we removed the backward branch and named it one-way recurrence. The ablation results are shown in Table \ref{tab:5}(c) and Table \ref{tab:5}(d).
It can be observed that the accuracy of one-way recurrence is much lower than that of bidirectional recurrence. In a one-way recurrence, all frames can only leverage the information of previous frames but cannot use information from subsequent frames. So, one-way recurrence may cause quite severe performance degradation.\\
\textbf{Flow-reuse strategy vs. naive flow estimation.}
To illustrate the effectiveness of the optical-flow-reuse strategy, we used two different optical flow estimation schemes as described in section~\ref{flow_method} for experiments.
When applying naive flow estimation between LR, we employ a pre-trained PWC-Net\cite{2017PWC} as the optical flow estimator and directly scale the estimated results to get the intermediate flow.
\begin{table}[t]
	\centering
	\caption{Ablation study on different modules.}\label{tab:5}
	\resizebox{0.90\linewidth}{!}{\begin{tabular}{l|llll}
		\hline
		Method                                                                 &(a)     &(b)      &(c)     &(d)        \\ \hline
		\begin{tabular}[c]{@{}l@{}}Bidirectional\\ recurrence\end{tabular}      &$\surd$& $\surd$&       &$\surd$         \\  \hline
		\begin{tabular}[c]{@{}l@{}}Intermediate flow\\ estimation\end{tabular} &       & $\surd$&$\surd$&$\surd$         \\  \hline
		\begin{tabular}[c]{@{}l@{}}Feature Refinement  \\ Module\end{tabular}   &       &       &$\surd$ &$\surd$         \\  \hline
		Vid4 (PSNR(dB))                                                                  & 26.14 & 26.68 &25.99   & 26.72 \\ \hline
		Vimeo (PSNR(dB))                                                                  & 35.00 & 35.41 &34.22   & 35.53 \\ \hline
	\end{tabular}}
\end{table}
We present the performance comparison of temporally interpolated frames on Vimeo-90K-T in Table \ref{tab:3} to more thoroughly explore the differences caused by the two optical flow estimation methods under various motion magnitudes. It can be seen that with the increase of motion magnitude, the advantage of intermediate flow estimation becomes more apparent, and naive flow estimation suffers from severe performance degradation since locally smooth assumption of optical may not hold in these cases. 
The visual comparison is given in Fig.~\ref{fig:5}, and one can see that there is severe distortion in the result of naive flow estimation. In contrast, the intermediate flow estimation can generate more visually pleasing results.
We may attribute the performance improvement to the direct estimation of intermediate optical flow. It can better accommodate non-linear motion between frames compared to linear motion estimation.\\ 
\textbf{Effectiveness of Feature Refinement Module.}
Since our network adopts a recurrent structure, when reconstructing a specific frame, we need to align the hidden states, which contain knowledge from other frames, to the current LR to obtain information gain.
In fact, it is feasible that we only use optical flow to perform 
alignment between frames. However, it is almost impossible for the optical flow to be completely accurate. The inaccurate optical flow estimation may be detrimental to restoring the current frame. The estimation error stems from two main reasons.  First, optical flow estimation itself is not completely accurate. Second, the proposed optical flow reuse strategy is also not entirely accurate. To alleviate the impact of inaccurate optical flow estimation, we propose the FRM to suppress those parts that do not look like the current frame.
In order to verify the effectiveness of feature refinement 
module, we remove it and compare the
performance on Vid4\cite{5995614} and Vimeo\cite{DBLP:journals/corr/abs-1711-09078}. The objective metrics are shown in (b) and (d) of the Table~\ref{tab:5}. 
The results validate that the Feature Refinement Module can further optimize the hidden feature.
We also display a visualization of the attention map in Fig.~\ref{fig:frm_vis} to explain the mechanism of FRM better. It can be seen that the parts with large movement (especially in some edge locations) have smaller values, while some flat areas have larger values. \\
\textbf{ Analysis of feature fusion space.}
Since we cannot access the ground truth of the synthetic intermediate frame, we must explore a reasonable 
way for estimating the intermediate states as accurately as possible.
The straightforward idea is to directly blend the warped
color frames in the image space to produce the intermediate frame. 
This approach is commonly used in image stitching \cite{2012Bibliography} and video extrapolation\cite{2019Extrapolation}.
However, image space fusion often easily leads to ghosting and checkerboard artifacts.
In order to avoid these negative effects, some work\cite{yi2021omniscient,2020TDAN} points out that fusion at the feature space will achieve better results. 
To this end, we explore the effect of feature fusion in different spaces,
namely:  image space fusion, feature space fusion, and hybrid-space fusion.
\begin{table}[t]
	\caption{Performance comparison (PSNR (dB) / SSIM ) of intermediate frames when applying two different flow methods under different motion magnitudes on Vimeo-90K-T\cite{DBLP:journals/corr/abs-1711-09078}. }
	\centering
	\resizebox{0.95\linewidth}{!}{\begin{tabular}{c|cl|cl|cl}
			\hline
			Flow estimation method  & \multicolumn{2}{c|}{Vimeo-Slow}   & \multicolumn{2}{c|}{Vimeo-Medium}  & \multicolumn{2}{c}{Vimeo-Fast}    \\ \hline
			Naive estimation        & \multicolumn{2}{c|}{32.49/0.9066} & \multicolumn{2}{c|}{33.45/0.9313}  & \multicolumn{2}{c}{32.97/0.9211}  \\ \hline
			Intermediate estimation & \multicolumn{2}{c|}{32.97/0.9133} & \multicolumn{2}{c|}{34.07/0.9315}  & \multicolumn{2}{c}{34.05/0.9850}  \\ \hline
			Performance Gain                    & \multicolumn{2}{c|}{0.48/0.0067}  & \multicolumn{2}{c|}{0.62/0.0070}   & \multicolumn{2}{c}{1.08/0.0102}   \\ \hline
	\end{tabular}}
	\label{tab:3}
\end{table}
When performing image space fusion, we use bidirectional optical flow to warp two adjacent
video frames and average the result of warped images.  Then we use a 1$\times$1 convolution to keep the 
channel dimension of the feature consistent with LR.  When applying feature space fusion,
we use bidirectional optical flow to warp adjacent hidden state and also average the results from both sides. When performing hybrid space fusion, we concatenate the results of image space fusion and feature space fusion in the channel dimension
and use 1$\times$1 convolution to process the result to ensure that the final channel
dimension is consistent with the LR inputs.
Quantitative results on the testset of Vimeo-90k-T are listed in Table \ref{tab6}.

\begin{table}[t]
	\centering
	\caption{Quantitative evaluation of fusion spaces on Vimeo-90k-T\cite{DBLP:journals/corr/abs-1711-09078}. }
	\label{tab6}
	\resizebox{0.85\linewidth}{!}{\begin{tabular}{l|l|l|c}
		\hline
		& Image-space & Feature-space     & \multicolumn{1}{l}{ Hybrid-space} \\ \hline
		PSNR(dB) &  \multicolumn{1}{c|}{35.11} & \multicolumn{1}{c|}{35.21} & 35.53                               \\ \hline
		SSIM &   \multicolumn{1}{c|}{0.9319}& \multicolumn{1}{c|}{0.9328} &  0.9359                               \\ \hline
	\end{tabular}}
\end{table}

It can be seen that hybrid space fusion can achieve the best results compared to the other two methods. This may be because the feature dimension is too low if only image-space fusion is used. Similarly, since the hidden state will inevitably lose information during the recurrence, only feature-space fusion cannot fully use all the features. So we apply hybrid space to achieve the best performance.  \\

\section{Conclusions}

\label{sec:concl}
This paper presents an Optical-Flow-Reuse-Based Bidirectional Recurrence Network (OFR-BRN) for space-time video super-resolution, which avoids redundant feature alignment operations and efficiently integrates long-range information using the continuously updated hidden state.  Specifically, our method directly generates bidirectional optical flow between the intermediate and adjacent frames. With the flow-reusing strategy, OFR-BRN can reorganize the optical flow for frame \& feature warping without superfluous motion information generation. Meanwhile, the proposed Bidirectional Recurrence Network reconstructs the low-resolution frames coarsely using bidirectional propagation to aggregate global temporal information. Then the Feature Refinement Module fuses the local spatial information according to the feature correlation, achieving further details refinement. Extensive experimental results on benchmark datasets demonstrate that OFR-BRN performs against the state-of-the-art ST-VSR algorithms in terms of accuracy and visual quality with high efficiency. 

Though the proposed method has achieved excellent results, some limitations still exist. First, Unlike TMNet\cite{xu2021temporal}, which can interpolate a time-arbitrary intermediate moment, our method can only infer the pre-defined intermediate moment determined by the flow estimator. Another limitation is that our approach fails to achieve satisfactory results with excessively large motions in some cases. This is also because we do not use any subsequent refinement\cite{DAIN} or complex synthetic structure\cite{2020Softmax} after feature interpolation to keep the network simple and efficient. In our future work, we will further consider how to deal with these cases and approximate more visually appealing results.

\bibliographystyle{IEEEtran}

\end{document}